# Image Fusion Using LEP Filtering and Bilinear Interpolation

**Haritha Raveendran[1], Deepa Thomas[2]**

*[1]P.G Scholar and [2]Assistant Professor Department of Computer Science, Musaliar College of Engineering and Technology, Pathanamthitta, India*

***Abstract*- Image Fusion is the process in which core information from a set of component images is merged to form a single image, which is more informative and complete than the component input images in quality and appearance. This paper presents a fast and effective image fusion method for creating high quality fused images by merging component images. In the proposed method, the input image is broken down to a two-scale image representation with a base layer having large scale variations in intensity, and a detail layer containing small scale details. Here fusion of the base and detail layers is implemented by means of a Local Edge preserving filtering based technique. The proposed method is an efficient image fusion technique in which the noise component is very low and quality of the resultant image is high so that it can be used for applications like medical image processing, requiring very accurate edge preserved images. Performance is tested by calculating PSNR and SSIM of images. The benefit of the proposed method is that it removes noise without altering the underlying structures of the image. This paper also presents an image zooming technique using bilinear interpolation in which a portion of the input image is cropped and bilinear interpolation is applied. Experimental results showed that the when PSNR value is calculated, the noise ratio is found to be very low for the resultant image portion.**

## I. INTRODUCTION

Image fusion is an efficient method for creating high quality images in a wide range of applications involving remote sensing and medical image processing. In image fusion, component images of the same scene can be merged into a single fused image. More elaborate information about the visual is obtained which is useful for human and machine perception. The requirements of a good image fusion method are the following. First, it should preserve most of the useful information of component images including the edges. Second, it would not produce image artifacts. Third, it should produce fast output and could be able to tolerate to conditions like noise in the input. A major drawback of guided filtering based image fusion is that it may over-smooth the resulting weights, which is not good for image fusion. To solve the problem mentioned above, an image fusion method with LEP (Local Edge Preserving) is proposed here. From experimental results it is found that the proposed LEP filtering based image fusion method gives much better performance when compared with other approaches. There are many advantages of the proposed image fusion method .It is a fast two-scale fusion method which does not rely heavily on a specific image decomposition method .Pixel saliency and spatial context for image fusion are achieved by a novel weight construction method. In this method the balance of pixel saliency and spatial consistency are obtained by adjusting the parameters of the LEP filter.

Bilinear interpolation is used to improve the quality of the image generated by using LEP filtering while zooming the image. The capability of interpolation to estimate the values of a continuous function from discrete samples is utilized here. This is implemented by determining the grey level value from the weighted average of four nearest pixels to the input coordinates, and assigning this value to the output coordinate

Experimental results showed that when image fusion with LEP filtering and bilinear interpolation is used, the parameters such as sharpness, naturalness, SSIM and PSNR values are preserved and the noise ratio is very low.

The existing image fusion technique based on guided filtering has a lot of drawbacks. When guided filtering is used for image fusion, the edges of resultant image are not well preserved, or in other words the pixels representing the edges are highly distorted. Second major drawback of existing fusion method is that it takes more time to produce image output. This is because of the time complexity associated with the guided filtering algorithm. While processing high resolution images, the existing system yields poor output. There is significant loss of quality of the original image. Third significant problem is the existence of artifacts in the fused image. Artifact is a noticeable distortion of image data while processing the image .This is found to be high in the existing system. Also when PSNR value is calculated, the noise ratio is found to be high and this indicates the presence significant noise component.

## II. PROPOSED METHOD

*Local Edge preserving Filter*

Local Edge preserving Filter can be used to generate edge preserving decomposition of an image .The local salient edges contained in the image are usually preserved in this method. The local energy function can be written in discrete form:

$$\sum_{i\epsilon w}(I_i - B_i)^2 + \frac{\alpha'}{|\nabla I_i|^{\beta}}|\nabla B_i|^2 \quad (1)$$

A Normalized Steepest Descent (NSD) method is used to minimize the above energy function to get a numerical solution, as shown in the Appendix. In order to create the solution, we suppose that *B* has a linear dependence with *I* in a local window, since pixels are highly correlated locally. Here we propose this local approximation of *B* as:

$$B_i = a_w I_i + I_i,\ i \in w, \quad (2)$$

Where $a_w$ and $b_w$ are constant coefficients represented in the window $\dot{\omega}$.

$$\sum_{i\epsilon w}(I_i - a_w I_i - b_w)^2 + \alpha' |\nabla I_i|^{2-\beta} . {a_w}^2 \quad (3)$$

The formula (3) is much like the cost function in except the coefficient $(\alpha' |\nabla I_i|^{2-\beta})$ between the two constraints. We will see later that our adaptive coefficient will preserve edges while that of other methods can't. Now the optimal problem becomes a parameter estimating problem. The minimum of (3) can be found by setting the partial derivative of each parameter to zero. This linear least squares' solution is:

$$\begin{cases} a_w = \dfrac{{\sigma_w}^2}{{\sigma_w}^2 + \frac{1}{N}\alpha' \sum_{i\epsilon w}|\nabla I_i|^{2-\beta}} \\ b_w = \bar{I}_w - a_w I_w, \end{cases} \quad (4)$$

where ${\sigma_w}^2$ is the variance of *I* in the window *w* and $\bar{I}_w$ is the mean of *I* in *w*. If $\alpha' = \beta = 1$, then $\frac{1}{N}.\alpha' \sum_{i\epsilon w}|\nabla I_i|^{2-\beta} = \frac{1}{N}.\sum_{i\epsilon w}|\nabla I_i|$ represents the average of the gradients in *w*. It can be easily deduced that $a_w$ is always less than 1, and the contrasts of the output of equation (2) will always be compressed. In other words, *B* can be considered as a smoothed version of *I* .Each window contains *N* pixels, and each pixel is involved in *N* windows. For every window, there is a set of $a_w$,$b_w$, and then, the filtered output $B_i$ of (2) has *N* different values. Weighted average method is applied to the values to retain correct results and diminish distorted ones. However, it is very difficult to figure out the weights, so we simply calculate the mean of all the *N* values of $B_i$. If a local window is identified by its central pixel, we change $a_k, b_k$ for $a_w, b_w$ and *k* denotes the central pixel's location. We get our LEP output as:

$$B_{i'} = \frac{1}{N}\sum_{k\in w}(a_k I_i + b_k) = \bar{a}_i I_i + \bar{b}_i \quad i \in \Omega \quad (5)$$

Where $\Omega$ represents the area of the image, and $\bar{a}_i$ is the average of the $a_k$ in the neighborhood window, and the same with $\bar{b}_i$ .We present our filtered result together with guided filter. We intentionally set the window radius (if it has) a large value for testing the edge preserving effect. Guided filter is not good at preserving edges. And our LEP seems to rank between us. It can preserve edges. This is just the feature of our LEP that local salient edges are well preserved in the filtered based layer. Our filter's advantage is the preserving of local edges. Another advantage of our LEP is that the algorithm's asymptotic time complexity is O(n), independent of the window size. Because the main operation is averaging of values, it can be implemented by box filters.

*Parameters for LEP*

The two parameters for LEP are $\alpha'$and β.These two parameters can determine filter's sensitivity and gradient. When the values of $\alpha'$ or $\beta$ is small, More gradients will be considered as salient edges. Otherwise, when $\alpha'$or $\beta$ is large, the resultant filtered output will be over smoothed (less gradients will be treated as salient edges). The visual clarity of the image will reduced with the increase of $\alpha'$ or $\beta$, while the details are kept with the decrease of $\alpha'$ or $\beta$. We find values for $\alpha' = 0.1$ and $\beta = 1$ to always produce good results, burring details while preserving salient edges.

### I. *Algorithm for Image Fusion using LEP Filtering*

First step in this approach is to create a two scale representation of input images by using an average filter. Then a weighted average method based on Local Edge Preserving filtering is employed for the fusion of the base and detail layers.

*A. Two-Scale Image Decomposition*

The source images are first decomposed into two-scale representations by means of average filtering. This is shown in figure 1. The base layer of source image may be represented as:

$$B_n = I_n * Z \quad (6)$$

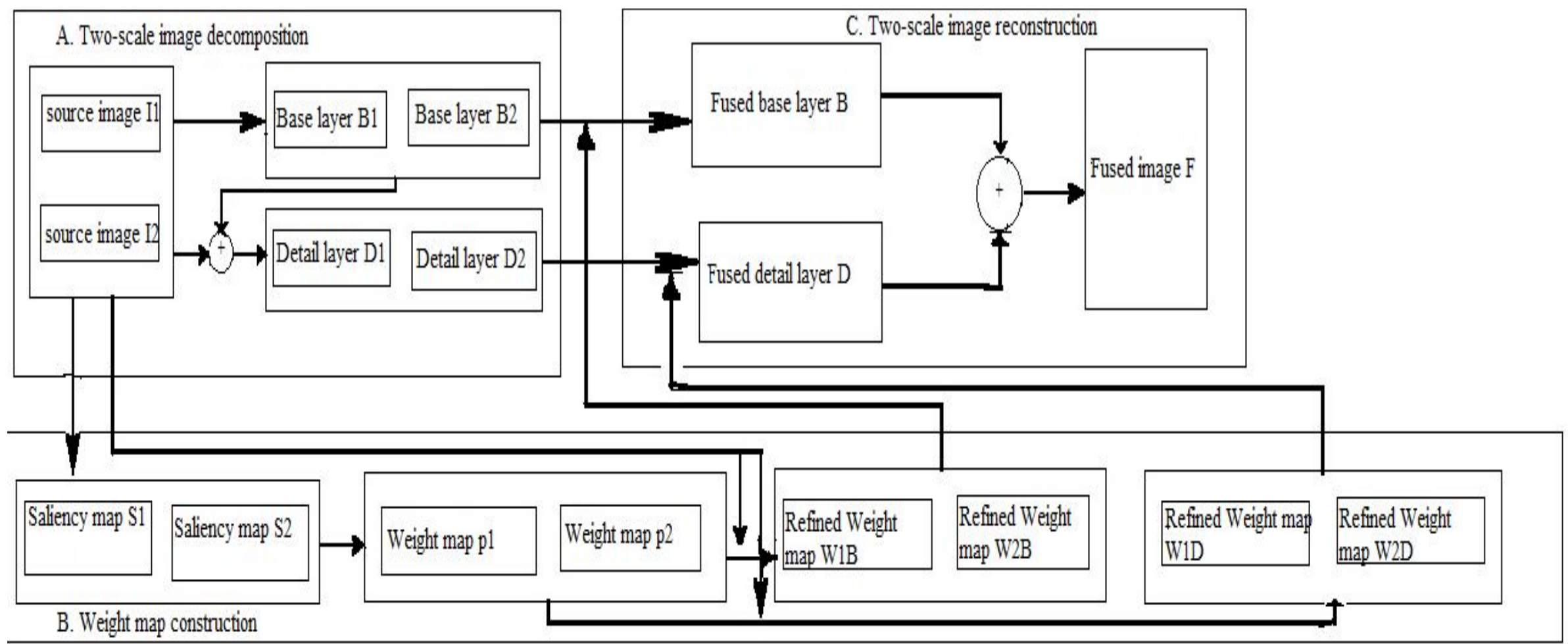

Fig.1: Flow chart of the Image fusion using LEP Filtering

Where $I_n$ is the *n*th source image, *Z* is the average filter, and the size of the average filter is usually set to 31×31. When the base layer is obtained, the detail layer can be easily obtained by subtracting the base layer from the source image. This may be represented as:

$$D_n = I_n - B_n \tag{7}$$

The objective of two-scale decomposition is to separate the input image into a layer representing large-scale variations in intensity and another one representing small scale details

*B. Weight Map Construction with LEP Filtering*

Weight Map Construction is the next step. As shown in Fig. 1, the weight map is created. Then we have to find the high-pass image $H_n$ by applying Laplacian filtering.

$$H_n = I_n * L \tag{8}$$

Where *L* is a 3 × 3 Laplacian filter. Then, the saliency maps $S_n$ is constructed using the local average of the absolute value of $H_n$.

$$S_n = |H_n| * g_{r_g,\sigma_g} \tag{9}$$

Where *g* represents a Gaussian low-pass filter of size $(2r_g + 1)$ $(2r_g + 1)$, and the parameters $r_g$ and $\sigma_g$ are set to 5. The measured saliency maps provide good characterization of the saliency level of detail information. Next process is the comparison of the saliency maps to determine the weight maps. This may be represented as follows:

$$P_n^{\,k}=1 \quad \text{if} \quad S_n^{\,k}=\max\,(S_1^{\,k}, S_2^{\,k}, \ldots, S_N^{\,k})$$

$$0 \quad \text{else} \tag{10}$$

Where *N* represents the number of source images, $S_n^{\,k}$ is the saliency value of the pixel *k* in the *n*th image. However, this method may produce artifacts to the fused image as the weight maps obtained by the above process are usually noisy and not aligned with object boundaries (see Fig. 1). This problem can be solved by using spatial consistency based approach. Spatial consistency means that if two adjacent pixels have similar brightness or color, they will usually tend to have similar weights. A commonly used spatial consistency based fusion approach is formulating an energy function, where the pixels saliencies are encoded in the function and edge aligned weights are enforced by regularization terms, e.g., a smoothness term. This energy function is then minimized globally to obtain the desired weight maps. The optimization based methods seems to be relatively inefficient. In this paper, we have proposed Local Edge Preserving ( LEP) image filtering. LEP filtering is performed on each weight map $P_n$ with the corresponding source image $I_n$ serving as the guidance image.

$$W_n^{\,B} = G_{r_1\epsilon_1}\ (P_n, I_n) \tag{11}$$

$$W_n^{\,D} = G_{r_2\epsilon_2}\ (P_n, I_n) \tag{12}$$

Where $r_1, \epsilon_1, r_2,$ and $\epsilon_2$ are the parameters of the LEP filter, $W_n^{\,B}$ and $W_n^{\,D}$ are the resulting weight maps of the base layer and the detail layer respectively. Finally, the values of the *N* weight maps are normalized such that they sum to one at each pixel *k*. The objective of the proposed weight construction method can be summarized as follows. According to (6), (8) and (9), it can be seen that if the local variance at a position *i* is very small which means that the pixel is in a flat area of the guidance image, then $a_k$ will become close to 0 and the filtering output *O* will equal to $\bar{P}_k$, i.e., the average of adjacent input pixels. In contrast, if the local variance of pixel *i* is very large which means that the pixel *i* is in an edge area, then $a_k$ will become far from zero. As demonstrated in, $\nabla O \approx \bar{a}\nabla I$ will become true, which means that only the weights in one side of the edge will be averaged. In both situations, those pixels with similar color or brightness tend to have similar weights. This is the core principle behind spatial consistency. The base layers look spatially smooth and thus the corresponding weights also should be spatially smooth. Otherwise, artificial edges may be produced. In contrast, sharp and edge-aligned weights are preferred for fusing the detail layers since details may be lost when the weights are over-smoothed. Therefore, the fusion of base layers requires a large filter size and a large blur degree while a small filter size and a small blur degree are suitable for the detail layers.

*C. Two-Scale Image Reconstruction*

Two-scale image reconstruction consists of the following two steps. First, weighted average method is used for the fusion of the base and detail layers of different source images.

$$\bar{B}=\sum_{n=1}^{N} W_n{}^B B_n \quad (13)$$

$$\bar{D}=\sum_{n=1}^{N} W_n{}^D D_n \quad (14)$$

Then, the fused base layer *B* and the fused detail layer *D* are combined to get the fused image *F* as represented below

$$F=\bar{B}+\bar{D} \quad (15)$$

## II. *Algorithm for Image Zooming with Bilinear Interpolation*

The Interpolation function may be represented mathematically as:

$$g(x) = \Sigma_k c_k u(\text{ distance }_k)$$

Here g() is the interpolation function, u() is the interpolation kernel, distance $_k$ indicates distance from the point under consideration, x, to a grid point, $x_k$, and $c_k$ are the coefficients of interpolation. The values of $c_k$ are selected to satisfy the condition $g(x_k) = f(x_k)$ for all $x_k$.

The proposed system applies Bilinear Interpolation to the resultant image generated by LEP filtering. This primarily aims to improve and enhance the quality of the fused image while zooming. The capability of interpolation to estimate the values of a continuous function from discrete samples is utilized here. This is implemented by determining the grey level value from the weighted average of the four closest pixels to the specified input coordinates, and assigning this value to the output coordinate. Four grid points are needed to evaluate the interpolation function in the case of Bilinear Interpolation. In the first step, two linear interpolations are applied in horizontal direction. Then one more linear interpolation is applied in the perpendicular direction.

The bilinear interpolation kernel may be represented as:

$$u(s) = \begin{cases} 0 & |s| > 1 \\ 1 - |s| & |s| < 1 \end{cases}$$

Where s represents distance between the point of interpolation and the grid point being considered. The interpolation coefficients $c_k = f(x_k)$.

## III. RESULTS

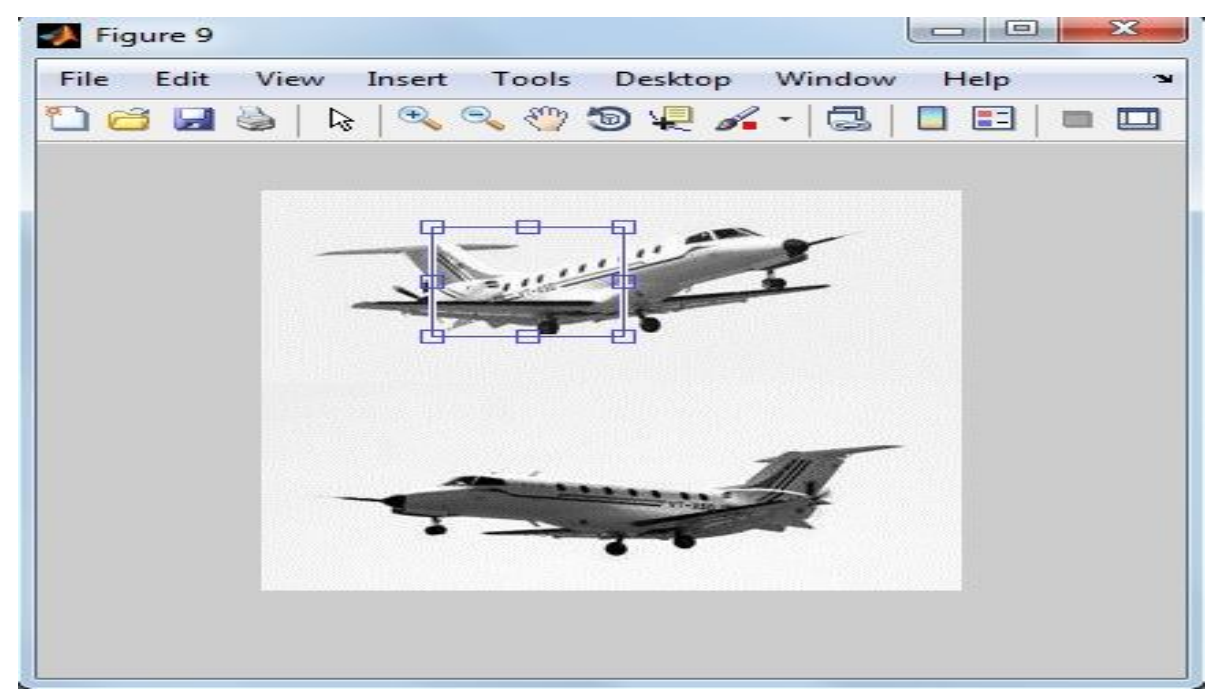

Fig.2: Bilinear interpolation of input image

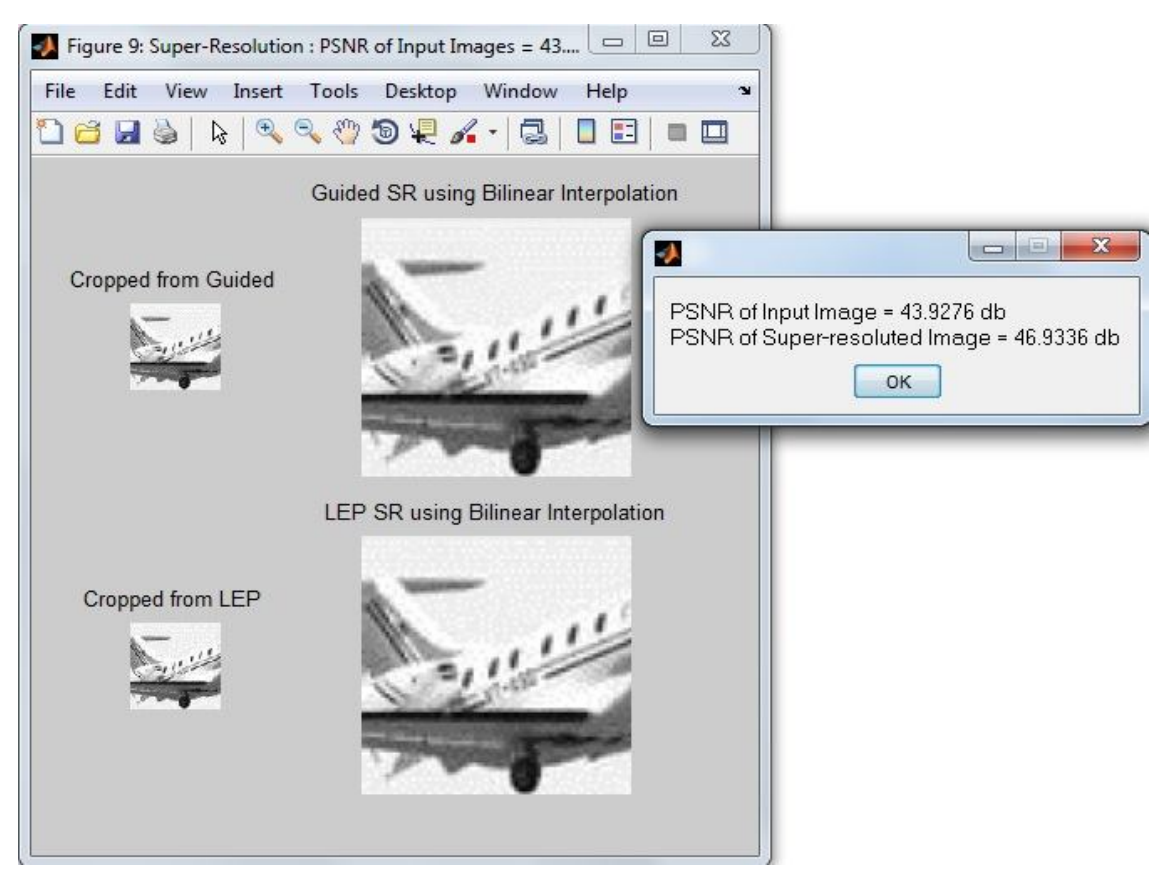

Fig.3: Comparison of PSNR value

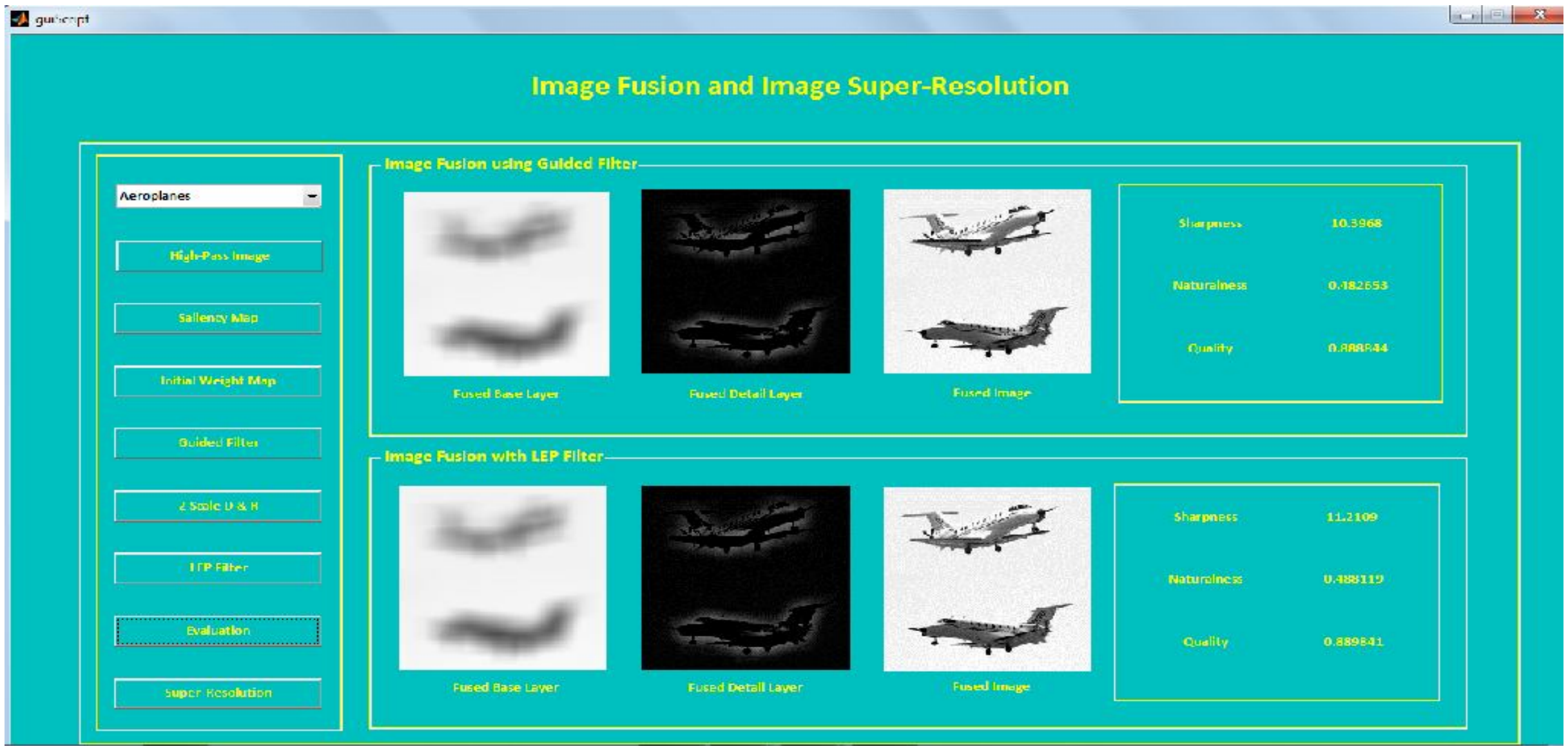

Fig.4: Evaluation of image fusion method

Table-1: Evaluation of image fusion methods

| Filters using Image Fusion | Sharpness | Naturalness | Quality |
|---|---|---|---|
| Guided Filter | 10.3968 | 0.482653 | 0.888844 |
| LEP Filter | 11.2109 | 0.488119 | 0.889841 |

## IV. CONCLUSION

In this paper we have proposed an image fusion method based on Local Edge Preserving filter. As the name indicates this algorithm can efficiently preserve the pixels representing the edges of resultant image. Most of the pixel information representing the edges of the image are retained and there is little loss of information. This method utilizes the average filter to get the two-scale representations, which is simple and more effective. More importantly, weight optimization is achieved by using the LEP filter in such a way as to make full use of the correspondence between adjacent pixels. Experiments show that when the proposed image fusion method is applied, the information contained in the component images is well preserved. The processing of medium to high resolution images using LEP filtering yields good output. The quality of the resultant image is also high in this method and the presence of artifacts is very rare. The application of Bilinear Interpolation technique further improves the quality of fused image when zooming is applied. The proposed method applies Image Zooming with Bilinear interpolation in addition to LEP filtering. When Bilinear interpolation is applied, the parameters such as sharpness, naturalness, SSIM and PSNR values are preserved and the noise ratio is found to be very low. The LEP filtering based image fusion with Bilinear interpolation is good for applications requiring high quality images like medical image processing. Experiments show that the proposed method using LEP filtering is many times faster than the existing guided filtering based image fusion. The overall performance of the proposed method can be improved in future researches by adaptively choosing the parameters of the LEP filter thereby obtaining supreme quality images even after zooming several times.